# Effective Damage Data Generation by Fusing Imagery with Human Knowledge Using Vision-Language Models*

Jie Wei
City College of New York
New York, NY, USA
jwei@ccny.cuny.edu

Erika Ardiles-Cruz*
Air Force Research Lab
Rome, NY, USA
erika.ardiles-cruz@us.af.mil

Aleksey Panasyuk
Air Force Research Lab
Rome, NY, USA
aleksey.panasyuk@us.af.mil

Erik Blasch
Air Force Research Lab
Arlington, VA, USA
erik.blasch.1@us.af.mil

*Abstract*—It is of crucial importance to assess damages promptly and accurately in humanitarian assistance and disaster response (HADR). Current deep learning approaches struggle to generalize effectively due to the imbalance of data classes, scarcity of moderate damage examples, and human inaccuracy in pixel labeling during HADR situations. To accommodate for these limitations and exploit state-of-the-art techniques in vision-language models (VLMs) to fuse imagery with human knowledge understanding, there is an opportunity to generate a diversified set of image-based damage data effectively. Our initial experimental results suggest encouraging data generation quality, which demonstrates an improvement in classifying scenes with different levels of structural damage to buildings, roads, and infrastructures.

## I. INTRODUCTION

Timely and accurate Damage Assessment (DA) [1, 2] is of crucial importance in humanitarian assistance and disaster response (HADR)[3] for understanding the impact of the events, allocating resources and personnel, guiding recovery efforts, and evaluating residual threats. Modern disasters can result in complex damage patterns that are difficult to discern from multimodal data (e.g., electro-optical, visual, and infrared data). Visual assessment, often from airborne or space-borne platforms (e.g., satellite imagery), remains a key DA input, especially for data fusion at the edge [4].

Recently, the xView2 dataset[5] represents a significant advancement in the HADR surveillance domain. The xView2 data contains a large set of annotated high-resolution satellite images for building damage assessment. In the xView2 dataset, damage scores for each building are carefully annotated and reviewed by expert analysts.

More than 850,000 building polygons from six different types of natural disasters, namely, hurricanes, volcanoes, fire, tsunamis, flooding, and earthquakes, around the world, are made available. The ground truths of the xView2 dataset were carefully reviewed by experts for accuracy and utility for the specific task of automated building localization and damage assessment with four different damage severity levels: no damage, minor, major, and total damage, making it the premier source of high-quality labeled imagery for DA and HADR[6].

The data is taken and annotated from the Joplin, MO tornado from 2011 [7]. It is estimated that 8000 buildings were damaged, with half of those being houses. The resulting damage influenced climatologists to create the "Waffle House Index," indicating that the index determines a 24 hrs/7-day-a-week opening status. The index was adopted by the Federal Emergency Management Agency (FEMA). The index lists three categories: green (has power and damage is minimal), yellow (power is provided by generator and supplies are low), and red (severe destruction and the building is closed).

Fig. 1 shows the locations of the xVIEW2 data from the Joplin hurricane. The red circles represent the most damage.

However, several inherent characteristics of xView2 and the broader challenge of HADR image analysis pose the following challenges:

1) *Class Imbalance*: The four damage levels are not balanced, with approximately 4.37% (4.48%), 0.54% (0.49%), 0.68% (0.57%), and 0.32% (0.31%) for no damage, minor, major, and total damage severity levels in the train (test) datasets, respectively. Most regions show no damage, while total damage regions are exceedingly rare.

2) *Occlusion and Distortion*: Clouds, shadows, and atmospheric distortions can obscure or alter building features, hindering accurate damage classification[8].

3) *Lighting Conditions*: Illumination variations (time of day, sun angle, overcast conditions) significantly alter image appearance and can confuse models[9].

4) *Seasonality*: Changes in vegetation, snow cover, and ground conditions across seasons affect building visibility and contextual information.

5) *Viewpoint, Scale, and Resolution Variability*: Imagery from different satellites or acquisition geometries exhibits variations in viewing angle (nadir vs. oblique), ground sampling distance, and overall resolution, impacting model generalization [10].

Given the aforementioned problems, despite the immense representational power, the deep learning models, without sufficient exposure to balanced training data of different damage levels, are unable to give rise to satisfactory generalization results. For instance, none of the winning teams in the XView2-based contests can achieve F1 scores above 0.8 [2]. It is practically impossible to cover all five possible scenarios to render them appropriate for deep net training. One way to partially resolve this problem is by resorting to deep learning-based data generation techniques, such as Generative Adversarial Networks (GAN)[11] and Diffusion Models (DM)[12], however, purely based learning from (unbalanced) examples and lack of high-level human-knowledge direction [13], and hence, the generated data from GAN and/or DM are generally of insufficient quality.

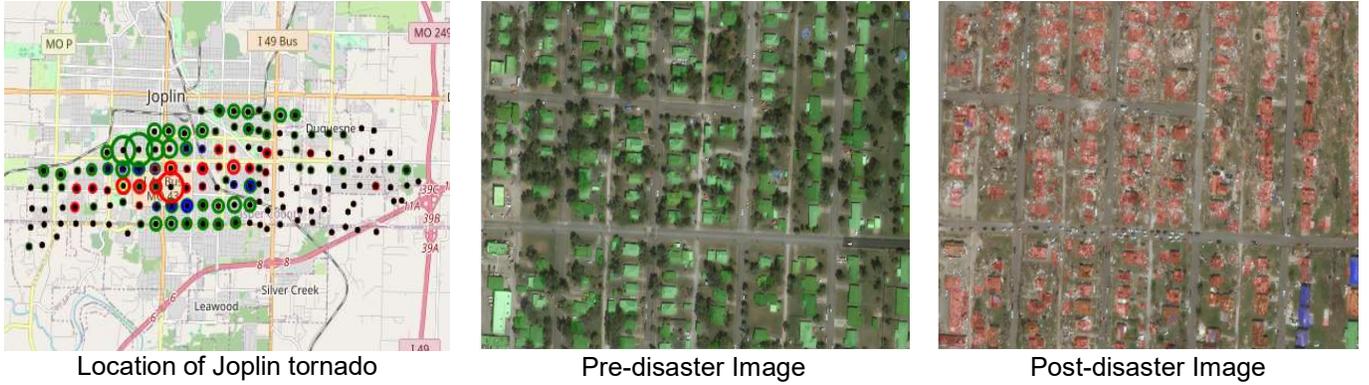

**Figure 1. Location of Joplin tornado and one sample pre- and post-disaster image.**

The breakthroughs initiated by the Large Language Models (LLMs), such as ChatGPT[14], Llama[15], and Gemini[16], revolutionized natural language processing. Afterward, the incorporation of vision deep nets, including YOLO [17], GAN, and DM, into the LLM further induced the vision-language models (VLM)[18], which can be exploited to fuse image-based visual information with natural-language-based human knowledge. VLMs can be used organically to generate damage data to address the challenges faced by HADR image data[19]. In this paper, we report our ongoing work to apply cutting-edge VLM techniques to generate image data of varying aspects for effective DA purposes.

## II. METHODOLOGY

In our previous work, (Contrastive Learning, Information Fusion, and Generative Adversarial Networks) CLIFGAN [2], the Pix2Pix conditional GAN architecture was used to generate images of different damage levels with subpar performance. One major problem was the lack of human guidance: the damage levels are purely based on the learning from given examples. For example, if the coverage of different scenarios is insufficient or lacking, the GAN cannot create data out of nothing or from too little information, as the neural network needs adequate examples to learn the innate statistics of the desired data, but it can only create statistically similar data to which it has already been exposed. Diffusion models (DM), despite different statistical methodologies, have the same trouble. VLM, by contrast, has the promise to resolve this issue because the high-level human or expert knowledge can be effectively encoded into the System/User prompts or reward/penalty, together with the input image(s) and some additional input images and corresponding desired outputs as few-shots examples, the unseen/novel images can be generated from VLM (in this paper, all results were generated by using Gemini 2.0/2.5-flash).

Traditional information fusion methods have long leveraged human experts for sensor control, collection, and assessment. For example, in Level 5 Information Fusion called "user refinement", the person assesses the data and object recognition (Level 1 information fusion) to determine the class labels from the data available. If the multimodal results are not accurate, the user would conduct sensor management, consult simulation or generated data, and utilize context in making a decision [20]. Level 2/3 information fusion is a situation assessment that can use and inform the human operator, such as in HADR situations; essentially operating as support to human-agent knowledge fusion [21].

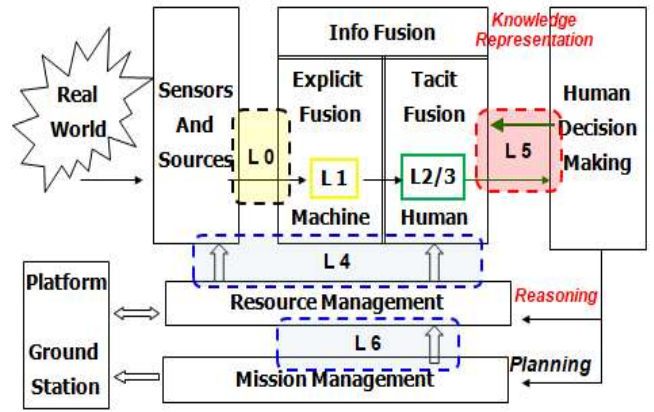

**Figure 2. Data Fusion Information Group Model.**

In our work, the following cutting-edge VLM techniques are used in our endeavor to generate damage data for various scenarios beyond those of traditional information fusion methods, as illustrated in Fig. 2:

1. *Prompt engineering*: the human/expert knowledge is carefully crafted in natural language form such that the system and user prompts guide the VLM in the generation of different damaged images[22]. In Fig. 3, the system prompt in our damage image generation is fed to Gemini as shown, where the official definition of the four different levels is given to explicitly guide the VLM to generate the damage images.

**System Prompt:**
   **The four damage levels are defined as below:**
   **No Damage**: Intact, operational structures. No visible impact from the disaster event.
   **Minor Damage**: Superficial damage, some missing non-structural elements (e.g., shingles, siding panels, broken windows), limited debris, structure remains sound.
   **Major Damage**: Significant structural compromise (e.g., partial roof/wall collapse, building visibly shifted off foundation, large sections missing, significant debris indicating structural failure). Habitation is questionable.
   **"Catastrophic/Total Damage"**: Building destroyed, largely collapsed, or structurally unsound beyond repair. Only foundations or extensive rubble may remain.

**Figure 3**. System prompt sent to the VLM dictating the specific four damage levels used in xView2 data.

2. *In-context learning*: instead of being purely based on natural language to encode human knowledge, we can also provide some sample images with the correct damage levels as ground truths, which are given to the VLM as desired input-output correspondence [23, 24], and hence multimodal human knowledge is accessible to the VLM for damage data generation.

3. *Chain-of-thought (COT) prompting*[25]: to improve the performance and yield a reliable, explainable, and transparent reasoning, an interactive process is invoked so that the expert and VLM can better communicate with each other. Elements include detailed explanations, system-level multimodal responses, and user prompts that are used together with several in-context input-output ground truths to improve the data generation quality.

4. *Active learning* with humans in the loop: if wrong damage data are created, which is the norm, especially in the initial phases, these incorrect ones should be singled out to either invoke a new COT procedure to correct it or memorized as in-context examples to be explicitly avoided [26].

5. *Retrieval Augmented Generation (RAG):* although the VLM was already exposed to a huge image and natural language dataset, for our highly specialized HADR applications, some unique knowledge and new expertise are unlikely to be available [27, 28]. To combat the need for data novelty and expertise accessibility, a local knowledge base is created and stored in a vector database as a local on-site expert, which can readily enrich the VLM with local knowledge. The local knowledge helps to avoid possible hallucinations, which can become an annoying phenomenon for most LLM/VLM.

6. *Low-Rank Adaptation* (LoRA) fine-tuning: besides RAG, for stable HADR specialized knowledge and definitions, we can fine-tune the VLM by decomposing the weight updates into smaller, trainable matrices using reinforcement learning (RL) techniques [29], which, combined with RAG, can yield more efficient and targeted damage data generation.

Besides the foregoing traditional VLM techniques, to achieve more precise damage content control, other vision foundation models such as grounding DINO [30], YOLO [31], the segment anything model (SAM) [32], Stable Fusion models [33], and instructPix2Pix [34], either as is or fine-tuned with xView2 dataset.

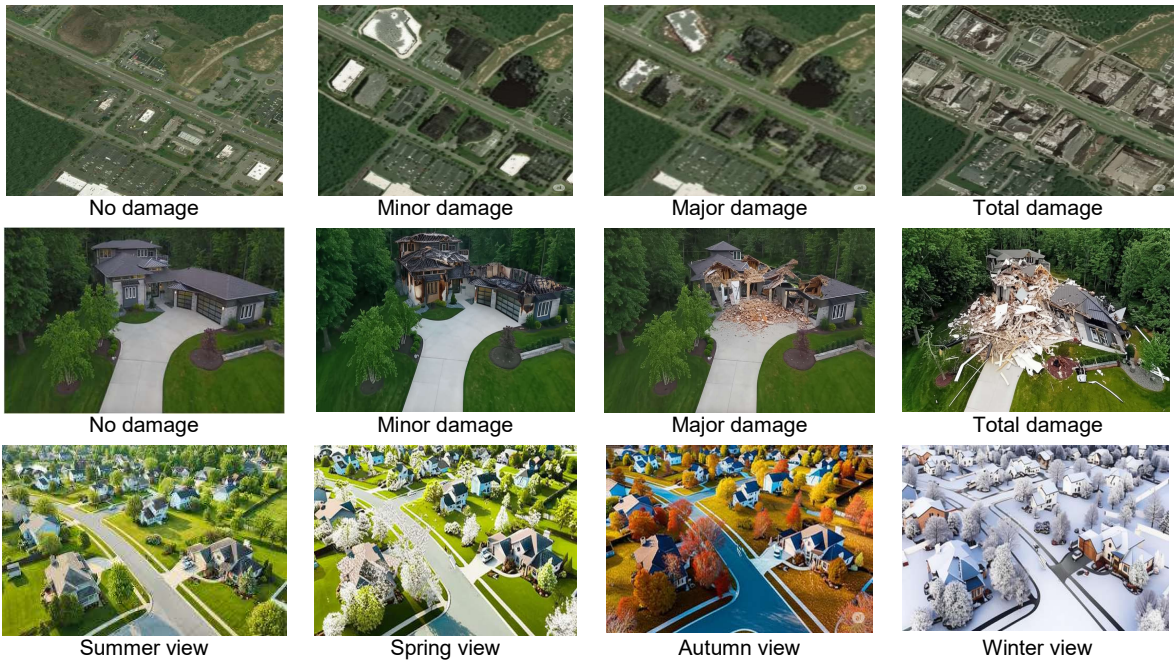

**Figure 4.** The VLM generated image data: Row 1: the Four damage levels for one sample xView2 image. Row 2: the four damage levels for a house; Row 3: the same neighborhood in different seasons: the original image is the summer, the spring, autumn, and winter views are generated.

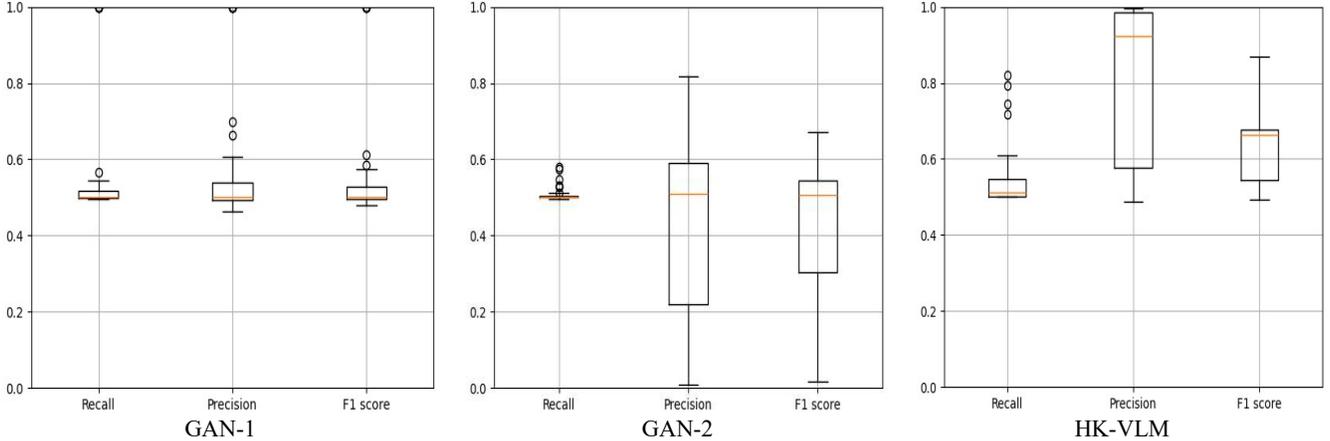

**Figure 5**. Boxplots of the CLIFGAN classification performance, namely, Recall, Precision, and F1 scores, over 50 images generated by GAN-1, GAN-2, and the HK-VLM.

### III. EXPERIMENTAL RESULTS OF THE DATA GENERATION BY FUSING NATURAL LANGUAGE AND VLM

As a qualitative illustration of the VLM techniques used for HADR purposes, Fig. 4 illustrates several sample images generated by Gemini. In Row 1, the images corresponding to the four damage levels of an xView2 image are depicted, which is visually more acceptable than the Pix2Pix GAN-generated ones in our previous work [2]. In Row 2, the four damage levels for the same house generated with VLMs satisfy the damage level definitions in Fig. 3. In Row 3, the same neighborhood shows seasonal terrain generated with VLMs, visually showing color differences and landscape change[35]. Illustrations show the power of VLMs to address the challenges of diversified data.

Besides qualitative observations shown in Fig. 4 as proof of the concepts of this new VLM CLIFGAN method, using the techniques described in the foregoing section affords more meaningful samples for the damage assessment classifier. Specifically, by combining the natural language based prompts with the *system prompts* shown in Fig. 3 and **user prompts** such as "*generate a satellite image with 2 small buildings with minor damages, each building must be very small with less than 10 percent of the image size, the other areas are forest and lakes*" and feeding them into **Gemini Flash 2.5**, multiple satellite images are created as shown in the first row of Fig. 4. The object-level masks, e.g., the buildings in the satellite images, are generated by using grounding Dino, YOLO, and the SAM, which is fine-tuned to better capture the 4-level damage exemplars as shown in Fig. 4. To gain insights into the effectiveness of this new human knowledge and VLM based image generation, denoted by human knowledge VLM (HK-VLM), we used the original image damage classifier, dubbed as the Contrastive Learning Image Fusion with GAN (CLIFGAN), trained over the xView2 dataset as reported in [2]. It is noted that the CLIFGAN was the 3[rd]-place award-winning damage data classification approach in the 2021 international overhead imagery Hackathon that focused on methods to provide novel change detection results.

To put in the context of the performance of HK-VLM, images generated by two different GAN approaches, GAN-1 and GAN-2, developed as part of the CLIFGAN system, more technical details of these two GANs were detailed in [2]. The classification performances achieved by the CLIFGAN system over GAN-1, GAN-2, and HK-VLM are reported in Table I, more statistical details are further illustrated in the three boxplots shown in Fig. 5.

**Table I. Classification performance (mean/median values) delivered by the CLIFGAN system over images generated by GAN-1, GAN-2, and HK-VLM, each with 50 images.**

|            | RECALL    | PRECISION | F1 SCORE  |
|------------|-----------|-----------|-----------|
| **GAN-1**  | 0.55/0.50 | 0.57/0.49 | 0.56/0.50 |
| **GAN-2**  | 0.51/0.50 | 0.43/0.50 | 0.46/0.51 |
| **HK-VLM** | **0.55/0.51** | **0.81/0.92** | **0.64/0.65** |

As summarized in Table I and Fig. 5, the average and median F1 scores for HK-VLM are 0.64 and 0.66, which are significantly better than the two heavily trained GAN models, both GAN-1 and GAN-2 took 12+ hrs each over our Nvidia RTX 3090 GPU, lagging far behind with values around 0.50. More interestingly, as reported in Table I of [2], CLIFGAN's overall damaged classification F1 score over the standard xView2 test dataset is 0.67, which is extremely similar to the value achieved by HK-VLM, 0.65, with statistically insignificant difference, which *suggests that the damage images generated by HK-VLM are indeed statistically similar to the true damage data used to train* and test the CLIFGAN, which demonstrated promising data generation quality. By contrast, the two rigorously trained GANs, as originally observed in our previous work [2], cannot give rise to such high-quality data.

## IV. Conclusion

It is of crucial importance to quickly and accurately identify damage levels for effective Humanitarian Assistance and Disaster Response (HADR) applications. However, the lack of or even the absence of targets of interest for certain damage levels and various operating conditions (e.g., lighting, seasonal, weather, and viewpoint) makes it hard to sufficiently train the deep nets to deliver the desired damage assessment performance. In this work, vision language models (VLMs) are explored as a means to fuse the image data and human knowledge to generate diversified damage data to assist HADR efforts. Cutting-edge VLM techniques such as prompt engineering, chain-of-thoughts, in-context learning, active learning, retrieval augmented generation, and low-rank adaptation-based fine-tuning are exploited for damage data generation.

Encouraging qualitative and quantitative results have been consistently observed via our initial tests by fusing human knowledge and VLM, which indicates that, compared with our own two different GAN architectures, this new approach by fusing human knowledge and VLM yields promising results. The work reported herein is merely the preliminary work this team has been taking on. More systematic studies will be performed to evaluate the effectiveness of these generated images as augmented data to train deep nets for HADR tasks. Two possible lines of attack will be examined next: 1) in this work, only CLIFGAN was used to determine the quantitative performance of the generated data; other objective methods, such as LLM as a judge [36], will be explored to shed more light on the quality of the generated images. 2) To emulate xBD datasets, in this work, we focused on small, controlled scenes. It will be of great interest to further generate large, geographically coherent areas with consistent damage patterns, which is often required for wide-area disaster assessment.

## Acknowledgment

The authors would like to gratefully acknowledge the support of AFRL/AFOSR FA9550-23-2-0002 and NSF IIS-2324052. Any opinions, findings, conclusions, or recommendations expressed in this material are those of the authors and do not necessarily reflect the views of the sponsors/organizations, the United States Air Force, or the US government.

## References

[1] E. Simoen, G. De Roeck, and G. Lombaert, "Dealing with uncertainty in model updating for damage assessment: A review," *Mechanical Systems and Signal Processing,* vol. 56, pp. 123-149, 2015.

[2] J. Wei *et al.*, "NIDA-CLIFGAN: Natural Infrastructure Damage Assessment through Efficient Classification Combining Contrastive Learning, Information Fusion and Generative Adversarial Networks," *Artificial Intelligence in Human Assistance and Disaster Response workshop, NeuraIPS'21, arXiv preprint arXiv:2110.14518,* p. 6, 2021.

[3] J. Wei, E. Blasch, E. Ardiles-Cruz, P. Morrone, and A. Aved, "Deep Learning Approach for Data and Computing Efficient Disaster Mitigation in Humanitarian Assistance and Disaster Response Applications," in *2022 IEEE International Humanitarian Technology Conference (IHTC)*, 2022: IEEE, pp. 79-85.

[4] A. Munir, E. Blasch, J. Kwon, J. Kong, and A. Aved, "Artificial intelligence and data fusion at the edge," *IEEE Aerospace and Electronic Systems Magazine,* vol. 36, no. 7, pp. 62-78, 2021.

[5] R. Gupta *et al.*, "Creating xBD: A dataset for assessing building damage from satellite imagery," in *Proceedings of the IEEE/CVF conference on computer vision and pattern recognition workshops*, 2019, pp. 10-17.

[6] E. Blasch and F. Darema, "Introduction to the DDDAS2022 Conference Infosymbiotics/Dynamic Data Driven Applications Systems," in *International Conference on Dynamic Data Driven Applications Systems*, 2022: Springer, pp. 3-13.

[7] https://en.wikipedia.org/wiki/Joplin_tornado (accessed.

[8] R. Szeliski, *Computer vision: algorithms and applications*. Springer Nature, 2022.

[9] M. Drew, J. Wei, and Z. N. Li, "Illumination-Invariance Image Retrieval and Video Segmentation," *Pattern Recognition,* vol. 32, no. 8, pp. 1369-1388, 1999.

[10] K. Murphy, *Machine Learning: A Probabilistic Perspective*. The MIT Press, 2012.

[11] I. Goodfellow *et al.*, "Generative adversarial nets," *Advances in neural information processing systems,* vol. 27, 2014.

[12] F.-A. Croitoru, V. Hondru, R. T. Ionescu, and M. Shah, "Diffusion models in vision: A survey," *IEEE Transactions on Pattern Analysis and Machine Intelligence,* 2023.

[13] E. Blasch *et al.*, "High level information fusion developments, issues, and grand challenges: Fusion 2010 panel discussion," in *2010 13th International Conference on Information Fusion*, 2010: IEEE, pp. 1-8.

[14] T. Wu *et al.*, "A brief overview of ChatGPT: The history, status quo and potential future development," *IEEE/CAA Journal of Automatica Sinica,* vol. 10, no. 5, pp. 1122-1136, 2023.

[15] H. Touvron *et al.*, "Llama 2: Open foundation and fine-tuned chat models," *arXiv preprint arXiv:2307.09288,* 2023.

[16] G. Team *et al.*, "Gemini: a family of highly capable multimodal models," *arXiv preprint arXiv:2312.11805,* 2023.

[17] P. Jiang, D. Ergu, F. Liu, Y. Cai, and B. Ma, "A Review of Yolo algorithm developments," *Procedia computer science,* vol. 199, pp. 1066-1073, 2022.

[18] G. Team *et al.*, "Gemini 1.5: Unlocking multimodal understanding across millions of tokens of context," *arXiv preprint arXiv:2403.05530,* 2024.

[19] E. Blasch, É. Bossé, and D. A. Lambert, *High-level information fusion management and systems design*. Artech House, 2012.

[20] E. P. Blasch, "Situation, impact, and user refinement," in *Signal Processing, Sensor Fusion, and Target Recognition XII*, 2003, vol. 5096: SPIE, pp. 463-472.

[21] D. Braines, A. Preece, C. Roberts, and E. Blasch, "Supporting agile user fusion analytics through human-agent knowledge fusion," in *2021 IEEE 24th International Conference on Information Fusion (FUSION)*, 2021: IEEE, pp. 1-8.

[22] L. Giray, "Prompt engineering with ChatGPT: a guide for academic writers," *Annals of biomedical engineering,* vol. 51, no. 12, pp. 2629-2633, 2023.

[23] Q. Dong *et al.*, "A survey on in-context learning," *arXiv preprint arXiv:2301.00234,* 2022.

[24] Y. Li, J. Wei, and C. Kamga, "Investigating Vision-Language Model for Point Cloud-based Vehicle Classification," *arXiv preprint arXiv:2504.08154,* 2025.

[25] J. Wei *et al.*, "Chain-of-thought prompting elicits reasoning in large language models," *Advances in neural information processing systems,* vol. 35, pp. 24824-24837, 2022.

[26] C. Brame, "Active learning," *Vanderbilt University Center for Teaching,* pp. 1-6, 2016.

[27] B. Ni *et al.*, "Towards Trustworthy Retrieval Augmented Generation for Large Language Models: A Survey," *arXiv preprint arXiv:2502.06872,* 2025.

[28] P. Lewis *et al.*, "Retrieval-augmented generation for knowledge-intensive nlp tasks," *Advances in neural information processing systems,* vol. 33, pp. 9459-9474, 2020.

[29] E. J. Hu *et al.*, "Lora: Low-rank adaptation of large language models," *ICLR,* vol. 1, no. 2, p. 3, 2022.

[30] S. Liu *et al.*, "Grounding dino: Marrying dino with grounded pre-training for open-set object detection," in *European Conference on Computer Vision*, 2024: Springer, pp. 38-55.

[31] J. Terven, D.-M. Córdova-Esparza, and J.-A. Romero-González, "A comprehensive review of yolo architectures in computer vision: From YOLOv1 to YOLOv8 and YOLO-NAS," *Machine Learning and Knowledge Extraction,* vol. 5, no. 4, pp. 1680-1716, 2023.

[32] A. Kirillov *et al.*, "Segment anything," *arXiv preprint arXiv:2304.02643,* 2023.

[33] A. Blattmann *et al.*, "Stable video diffusion: Scaling latent video diffusion models to large datasets," *arXiv preprint arXiv:2311.15127,* 2023.

[34] T. Brooks, A. Holynski, and A. A. Efros, "Instructpix2pix: Learning to follow image editing instructions," in *Proceedings of the IEEE/CVF conference on computer vision and pattern recognition*, 2023, pp. 18392-18402.

[35] Y. Zheng, E. Blasch, and Z. Liu, *Multispectral image fusion and colorization*. SPIE press Bellingham, Washington, 2018.

[36] J. Gu *et al.*, "A survey on llm-as-a-judge," *arXiv preprint arXiv:2411.15594,* 2024.